\documentclass{article} 
\usepackage{iclr2024_conference,times}

\usepackage{amsmath,amsfonts,bm}

\def\eqref#1{equation~\ref{#1}}

\def\1{\bm{1}}

\DeclareMathAlphabet{\mathsfit}{\encodingdefault}{\sfdefault}{m}{sl}
\SetMathAlphabet{\mathsfit}{bold}{\encodingdefault}{\sfdefault}{bx}{n}

\usepackage{url}

\usepackage{bbding}
\makeatletter
  \newcommand\figcaption{\def\@captype{figure}\caption}
  \newcommand\tabcaption{\def\@captype{table}\caption}
\makeatother

\usepackage{booktabs}       
\usepackage{amsfonts}       
\usepackage{nicefrac}       
\usepackage{microtype}      
\usepackage{xcolor}         
\usepackage{color, colortbl}
\definecolor{citecolor}{HTML}{2980b9}
\definecolor{linkcolor}{HTML}{c0392b}
\newcommand\blfootnote[1]{%
  \begingroup
  \renewcommand\thefootnote{}\footnote{#1}%
  \addtocounter{footnote}{-1}%
  \endgroup
}
\usepackage{amsmath}
\usepackage{amssymb}
\usepackage{graphicx}

\usepackage{multirow}
\usepackage{makecell}
\usepackage{caption}

\usepackage{adjustbox}

\usepackage[hidelinks,breaklinks=true,colorlinks,bookmarks=false,citecolor=citecolor,linkcolor=linkcolor]{hyperref}

\title{Personalize Segment Anything Model with\\One Shot}
\author{Renrui Zhang$^{1,2}$, Zhengkai Jiang$^{3*}$, Ziyu Guo$^{2*}$, Shilin Yan$^{2}$, Junting Pan$^{1}$, Xianzheng Ma$^{2}$\\\textbf{Hao Dong$^{4}$, Yu Qiao$^{2}$, Peng Gao$^{2}$, Hongsheng Li$^{1\dagger}$}\vspace{0.3cm}\\
  $^1$CUHK MMLab\quad 
  $^2$Shanghai Artificial Intelligence Laboratory\\
  $^3$Institute of Automation, Chinese Academy of Sciences\\
  $^4$CFCS, School of CS, Peking University\vspace{0.1cm}\\
\texttt{\{zhangrenrui, gaopeng, guoziyu, qiaoyu\}@pjlab.org.cn},\\
\texttt{kaikaijiang.jzk@gmail.com}\ \
\texttt{hsli@ee.cuhk.edu.hk}
}

\iclrfinalcopy 
\begin{document}

\maketitle
\blfootnote{*\ Equal contribution.\quad $\dagger$\ Corresponding author.}

\begin{abstract}
Driven by large-data pre-training, Segment Anything Model (SAM) has been demonstrated as a powerful promptable framework, revolutionizing the segmentation field. 
    Despite the generality, customizing SAM for specific visual concepts without man-powered prompting is under-explored, e.g., automatically segmenting your pet dog in numerous images.
    In this paper, we introduce a training-free \textbf{Per}sonalization approach for SAM, termed \textbf{PerSAM}. 
    Given only one-shot data, i.e., a single image with a reference mask, we first obtain a positive-negative location prior for the target concept in new images. Then, aided by target visual semantics, we empower SAM for personalized object segmentation via two proposed techniques: target-guided attention and target-semantic prompting. In this way, we can effectively customize the general-purpose SAM for private use without any training. 
    To further alleviate the ambiguity of segmentation scales, we present an efficient one-shot fine-tuning variant, \textbf{PerSAM-F}. Freezing the entire SAM, we introduce a scale-aware fine-tuning to aggregate multi-scale masks, which only tunes \textit{\textbf{2 parameters}} within \textit{\textbf{10 seconds}} for improved performance. 
    To demonstrate our efficacy, we construct a new dataset, PerSeg, for the evaluation of personalized object segmentation, and also test our methods on various one-shot image and video segmentation benchmarks.
    Besides, we propose to leverage PerSAM to improve DreamBooth for personalized text-to-image synthesis. By mitigating the disturbance of training-set backgrounds, our approach showcases better target appearance generation and higher fidelity to the input text prompt. Code is released at \url{https://github.com/ZrrSkywalker/Personalize-SAM}.
\end{abstract}

\begin{figure*}[h]
  \centering
\includegraphics[width=\textwidth]{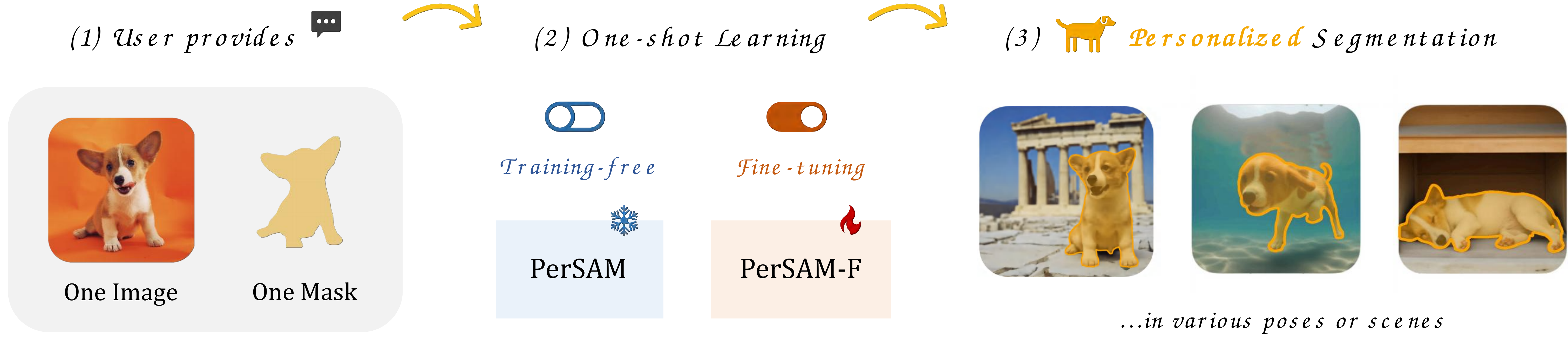}
   \caption{\textbf{Personalization of Segment Anything Model.} We customize Segment Anything Model (SAM)~\citep{kirillov2023segment} for specific visual concepts, e.g., your pet dog. With only one-shot data, we introduce two efficient solutions: a training-free PerSAM, and a fine-tuning PerSAM-F.}
    \label{fig1}
\end{figure*}

\section{Introduction}

Foundations models in vision~\citep{li2022uni,zou2023segment,wang2022images}, language~\citep{brown2020language,touvron2023llama,radford2019language}, and multi-modality~\citep{radford2021learning,jia2021scaling,li2023blip} have gained unprecedented prevalence, attributed to the availability of large-scale datasets and computational resources. They demonstrate extraordinary generalization capacity in zero-shot scenarios, and display versatile interactivity incorporating human feedback. Inspired by this, Segment Anything~\citep{kirillov2023segment} develops a delicate data engine for collecting 11M image-mask data, and subsequently trains a segmentation foundation model, known as SAM. It defines a novel promptable segmentation framework, i.e., taking as input a handcrafted prompt and returning the expected mask, which allows for segmenting any objects in visual contexts.

However, SAM inherently loses the capability to segment specific visual concepts. Imagine intending to crop your lovely pet dog in a thick photo album, or find the missing clock from a picture of your bedroom.
Utilizing the vanilla SAM would be highly labor-intensive and time-consuming. For each image, you must precisely find the target object within complicated contexts, and then activate SAM with a proper prompt for segmentation. Considering this, we ask: \textit{Can we personalize SAM to automatically segment user-designated visual concepts in a simple and efficient manner?}


\begin{figure*}[t!]
\vspace{-0.2cm}
\begin{minipage}[t]{0.48\textwidth}
\includegraphics[width=\textwidth]{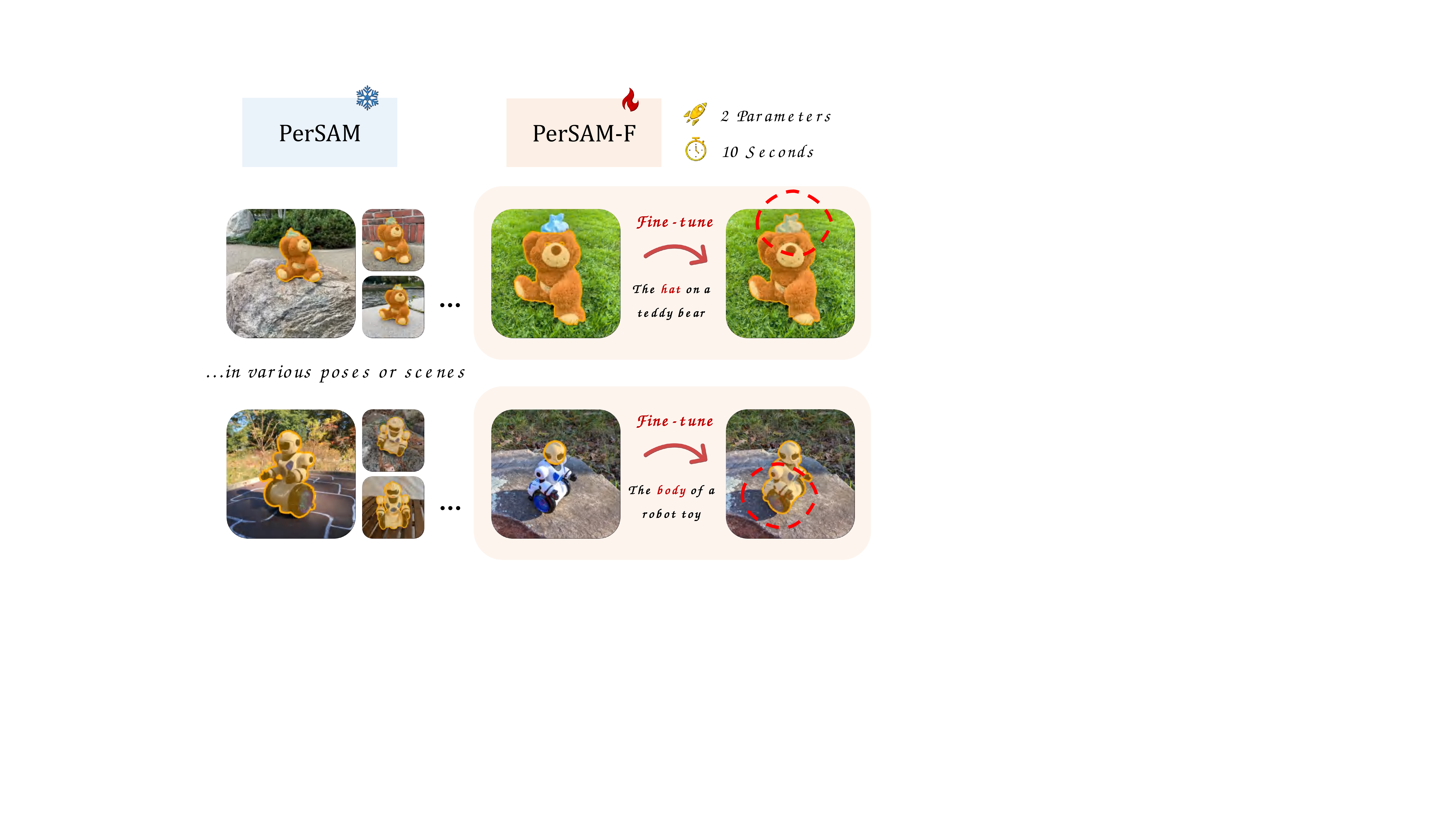}
\figcaption{\textbf{Personalized Segmentation Examples.} Our PerSAM (Left) can segment personal objects in any context with favorable performance, and PerSAM-F (right) further alleviates the ambiguity issue by scale-aware fine-tuning.}
\label{fig2}
\end{minipage}
\hspace{0.1in}
\vspace{-0.2cm}
\begin{minipage}[t]{0.48\textwidth}
\includegraphics[width=\textwidth]{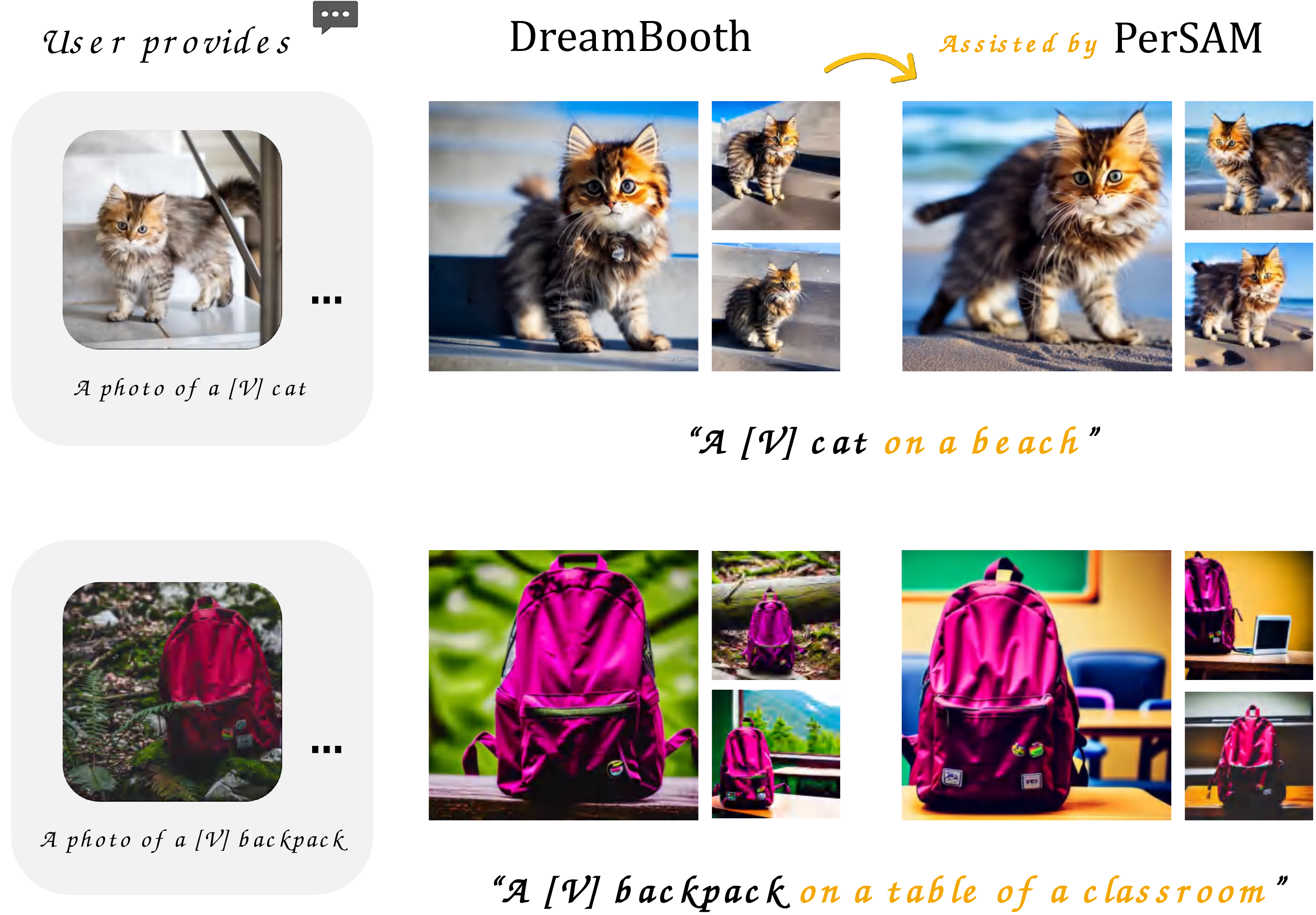}
\figcaption{\textbf{Improving DreamBooth~\citep{ruiz2022dreambooth} with PerSAM.} By mitigating the disturbance of backgrounds during training, our approach can help to achieve higher-quality personalized text-to-image generation.}
\label{fig_db1}
\end{minipage}
\end{figure*}

To this end, we introduce \textbf{PerSAM}, a training-free personalization approach for Segment Anything Model. As shown in Figure~\ref{fig1}, our method efficiently customizes SAM using only one-shot data, i.e., a user-provided reference image and a rough mask of the personal concept. Specifically, we first obtain a location confidence map for the target object in the test image by feature similarities, which considers the appearance of every foreground pixel. According to confidence scores, two points are selected as the positive-negative location prior, which are finally encoded as prompt tokens and fed into SAM's decoder for segmentation. Within the decoder, we propose
to inject visual semantics of the target object to unleash SAM's personalized segmentation power with two techniques:

\begin{itemize}
   \item \textbf{Target-guided Attention.}
    We guide every token-to-image cross-attention layer in SAM's decoder by the location confidence map. This explicitly compels the prompt tokens to mainly concentrate on foreground target regions for intensive feature aggregation.

    \item \textbf{Target-semantic Prompting.}
   To explicitly provide SAM with high-level target semantics, we fuse the original prompt tokens with the embedding of the target object, which provides the low-level positional prompt with additional visual cues for personalized segmentation.

\end{itemize}

With the aforementioned designs, along with a cascaded post-refinement, PerSAM exhibits favorable personalized segmentation performance for unique subjects in a variety of poses or scenes. Notably, our approach can cope well with scenarios that require segmenting one object among multiple similar ones, simultaneously segmenting several identical objects in the same image, or tracking different objects along a video.
Nevertheless, as shown in Figure~\ref{fig2}, there might be occasional failure cases, where the object comprises visually distinct subparts or hierarchical structures to be segmented, e.g., the hat on top of a teddy bear, or the head of a robot toy.
Such ambiguity casts a challenge for PerSAM in determining the appropriate scale of mask as output, since both the local part and the global shape can be regarded as valid masks by SAM.

To alleviate this issue, we further propose a fine-tuning variant of our approach, \textbf{PerSAM-F}. We freeze the entire SAM to preserve its versatile pre-trained knowledge, and only fine-tune \textbf{\textit{2 parameters}} within \textbf{\textit{10 seconds}} on a single A100 GPU. In detail, we enable SAM to produce several potential segmentation results of different mask scales. To adaptively select the best scale for varying objects, we employ a learnable relative weight for each mask scale, and conduct a weighted summation as the final output.
By such efficient scale-aware training, PerSAM-F avoids over-fitting on the one-shot data and exhibits better segmentation accuracy shown in Figure~\ref{fig2} (Right).

Moreover, we observe that our approach can also assist DreamBooth~\citep{ruiz2022dreambooth} to better fine-tune diffusion models for personalized text-to-image generation, as shown in Figure~\ref{fig_db1}. Given a few images containing a specific visual concept, e.g., your pet cat or backpack, DreamBooth learns to convert these images into an identifier [V] in the word embedding space, which, however, can simultaneously include the background information, e.g., stairs or the forest. This would override the newly prompted backgrounds, and disturb the target appearance generation. Therefore, we propose to leverage PerSAM to segment the target object within training images, and only supervise DreamBooth by the foreground area, enabling text-to-image synthesis with higher quality.

We summarize the contributions of our paper as follows:

\begin{itemize}
   \item \textbf{Personalized Object Segmentation.}
   We first investigate how to customize a general-purpose segmentation model (SAM) into personalized scenarios with minimal expense. To this end, we introduce two efficient and effective methods, along with a new segmentation dataset, PerSeg, for the evaluation of personalized object segmentation.
   
   \item \textbf{PerSAM and PerSAM-F.}
   In PerSAM, we propose three training-free techniques to guide SAM by the high-level semantics of target objects. In PerSAM-F, we design a scale-aware fine-tuning with 2 parameters in 10 seconds to well alleviate the mask ambiguity issue.

    \item 
    Our approach achieves competitive results on various tasks, including the PerSeg benchmark, one-shot part and semantic segmentation, and video object segmentation. In addition, PerSAM can enhance DreamBooth for better personalized text-to-image synthesis.
\end{itemize}

\section{Related Work}
\label{related}

\paragraph{Foundation Models.} 
With powerful generalization capacity, pre-trained foundation models can be adapted for various downstream scenarios and attain promising performance. In natural language processing, BERT~\citep{devlin2018bert,lu2019vilbert}, GPT series~\citep{brown2020language,OpenAI2023GPT4TR,Radford2018ImprovingLU,radford2019language}, and LLaMA~\citep{zhang2023llama} have demonstrated remarkable in-context learning abilities, and can be transferred to new tasks by domain-specific prompts. Similarly, CLIP~\citep{radford2021learning} and ALIGN~\citep{jia2021scaling}, which conduct contrastive learning on image-text pairs, exhibit exceptional accuracy in zero-shot visual recognition.
Painter~\citep{wang2022images} introduces a vision model that unifies network architectures and in-context prompts to accomplish diverse vision tasks, without downstream fine-tuning.
CaFo~\citep{zhang2023prompt} cascades different foundation models and collaborates their pre-trained knowledge for robust low-data image classification.
SAM~\citep{kirillov2023segment} presents a foundation model for image segmentation, which is pre-trained by 1 billion masks and conducts prompt-based segmentation. 
There are some concurrent works extending SAM for high-quality segmentation~\citep{ke2023segment}, faster inference speed~\citep{zhao2023fast,zhang2023faster}, all-purpose matching~\citep{liu2023matcher}, 3D reconstruction~\citep{cen2023segment}, object tracking~\citep{yang2023track}, medical~\citep{ma2023segment,huang2023segment} image processing.
From another perspective, we propose to personalize the segmentation foundation model, i.e., SAM, for specific visual concepts, which adapts a generalist into a specialist with only one shot.
Our method can also assist the personalization of text-to-image foundation models, i.e., Stable Diffusion~\citep{rombach2022high} and Imagen~\citep{saharia2022photorealistic}, which improves the generation quality by segmenting the foreground target objects from the background disturbance.

\paragraph{Large Models in Segmentation.} 
As a fundamental task in computer vision, segmentation~\citep{long2015fully, jiang2022prototypical, zhao2017pyramid, xu2021learning, jiang2023stc,lin2022frozen} requires a pixel-level comprehension of a image. Various segmentation-related tasks have been explored, such as semantic segmentation, classifying each pixel into a predefined set of classes~\citep{badrinarayanan2017segnet, chen2017deeplab, zheng2021rethinking, cheng2022masked, xie2021segformer, song2020rethinking}; instance segmentation, focusing on the identification of individual object instances~\citep{he2017mask,wang2020solov2,tian2020conditional}; panoptic segmentation, assigning both class labels and instance identification~\citep{kirillov2019panoptic,li2019attention}; and interactive segmentation, involving human intervention for refinement~\citep{hao2021edgeflow, chen2021conditional}.
Recently, inspired by language foundation models~\citep{zhang2023llama,brown2020language}, several concurrent works have proposed large-scale vision models for image segmentation. They are pre-trained by extensive mask data and exhibit strong generalization capabilities on numerous image distributions.
Segment Anything Model (SAM)~\citep{kirillov2023segment} utilizes a data engine with model-in-the-loop annotation to learn a promptable segmentation framework, which generalizes to downstream scenarios in a zero-shot manner.
Painter~\citep{wang2022images} and SegGPT~\citep{wang2023seggpt} introduce a robust in-context learning paradigm and can segment any images by a given image-mask prompt.
SEEM~\citep{zou2023segment} further presents a general segmentation model prompted by multi-modal references, e.g., language and audio, incorporating versatile semantic knowledge.
In this study, we introduce a new task termed personalized object segmentation, and annotate a new dataset PerSeg for evaluation.
Instead of developing large segmentation models, our goal is to personalize them to segment user-provided objects in any poses or scenes. 
We propose two approaches, PerSAM and PerSAM-F, which efficiently customize SAM for personalized segmentation.

\paragraph{Parameter-efficient Fine-tuning.} Directly tuning the entire foundation models on downstream tasks can be computationally expensive and memory-intensive, posing challenges for resource-constrained applications. To address this issue, recent works have focused on developing parameter-efficient methods~\citep{sung2022vl,he2022towards,rebuffi2017learning,qin2021learning} to freeze the weights of foundation models and append small-scale modules for fine-tuning. Prompt tuning~\citep{lester2021power, zhou2022learning, jia2022visual,liu2021p-tuning} suggests using learnable soft prompts alongside frozen models to perform specific downstream tasks, achieving more competitive performance with scale and robust domain transfer compared to full model tuning. Low-Rank Adaption (LoRA)~\citep{hu2021lora,stable-diffusion-lora,zhang2023adaptive,hedegaard2022structured} injects trainable rank decomposition matrices concurrently to each pre-trained weight, which significantly reduces the number of learnable parameters required for downstream tasks. Adapters~\citep{houlsby2019parameter,pfeiffer2020adapterfusion,lin2020exploring, chen2022vision} are designed to be inserted between layers of the original transformer, introducing lightweight MLPs for feature transformation. 
Different from existing works,
we adopt a more efficient adaption method delicately designed for SAM, i.e., the scale-aware fine-tuning of PerSAM-F with only 2 parameters and 10 seconds. This effectively avoids the over-fitting issue on one-shot data, and alleviates the ambiguity of segmentation scale with superior performance.

\section{Method}

In Section~\ref{s3.1}, we first briefly revisit Segment Anything Model (SAM)~\citep{kirillov2023segment}, and introduce the task definition for personalized object segmentation. Then, we illustrate the methodology of our PerSAM and PerSAM-F in Section~\ref{s3.2} and~\ref{s3.3}, respectively. Finally, we utilize our approach to assist DreamBooth~\citep{ruiz2022dreambooth} for better text-to-image generation in Section~\ref{s3.4}.

\subsection{Personalized Object Segmentation}
\label{s3.1}

\paragraph{A Revisit of Segment Anything.} 
SAM consists of three components, a prompt encoder, an image encoder, and a lightweight mask decoder, respectively denoted as $\operatorname{Enc}_P$, $\operatorname{Enc}_I$, and $\operatorname{Dec}_M$. 
As a promptable framework, SAM takes as input an image $I$, and a set of prompts $P$, which can be a point, a box, or a coarse mask. Specifically, SAM first utilizes $\operatorname{Enc}_I$ to obtain the input image feature, and adopts $\operatorname{Enc}_P$ to encode the human-given prompts of a length $k$ into prompt tokens as
\begin{align}
    F_I = \operatorname{Enc}_I(I),\ \ \ T_P = \operatorname{Enc}_P(P),
\end{align}
where $F_I \in \mathbb{R}^{h\times w\times c}$ and $T_P \in \mathbb{R}^{k\times c}$, with $h, w$ denoting the resolution of the image feature map and $c$ denoting the feature dimension.
After that, the encoded image and prompts are fed into the decoder $\operatorname{Dec}_M$ for attention-based feature interaction. SAM constructs the input tokens of the decoder by concatenating several learnable mask tokens $T_M$ as prefixes to the prompt tokens $T_P$. These mask tokens are responsible for generating the mask output, formulated as
\begin{align}
    M = \operatorname{Dec}_M\Big(F_I,\ \operatorname{Concat}(T_M, T_P)\Big),
\label{input_tokens}
\end{align}
where $M$ denotes the final segmentation mask predicted by SAM.

\paragraph{Task Definition.} 
Although SAM is generalized enough for any object by prompting, it lacks the ability to automatically segment specific subject instances. Considering this, we define a new task for personalized object segmentation. The user provides only a single reference image, and a mask indicating the target visual concept. The given mask can either be an accurate segmentation, or a rough sketch drawn on-the-fly. Our goal is to customize SAM to segment the designated object within new images or videos, without additional human prompting. 
For evaluation, we annotate a new dataset for personalized segmentation, named PerSeg. The raw images are collected from the works for subject-driven diffusion models~\citep{gal2022image,ruiz2022dreambooth,kumari2022multi}, containing various categories of visual concepts in different poses or scenes. In this paper, we propose two efficient solutions for this task, which we specifically illustrate as follows.

\begin{figure*}[t!]
\vspace{-0.2cm}
\begin{minipage}[t]{0.47\textwidth}
\includegraphics[width=\textwidth]{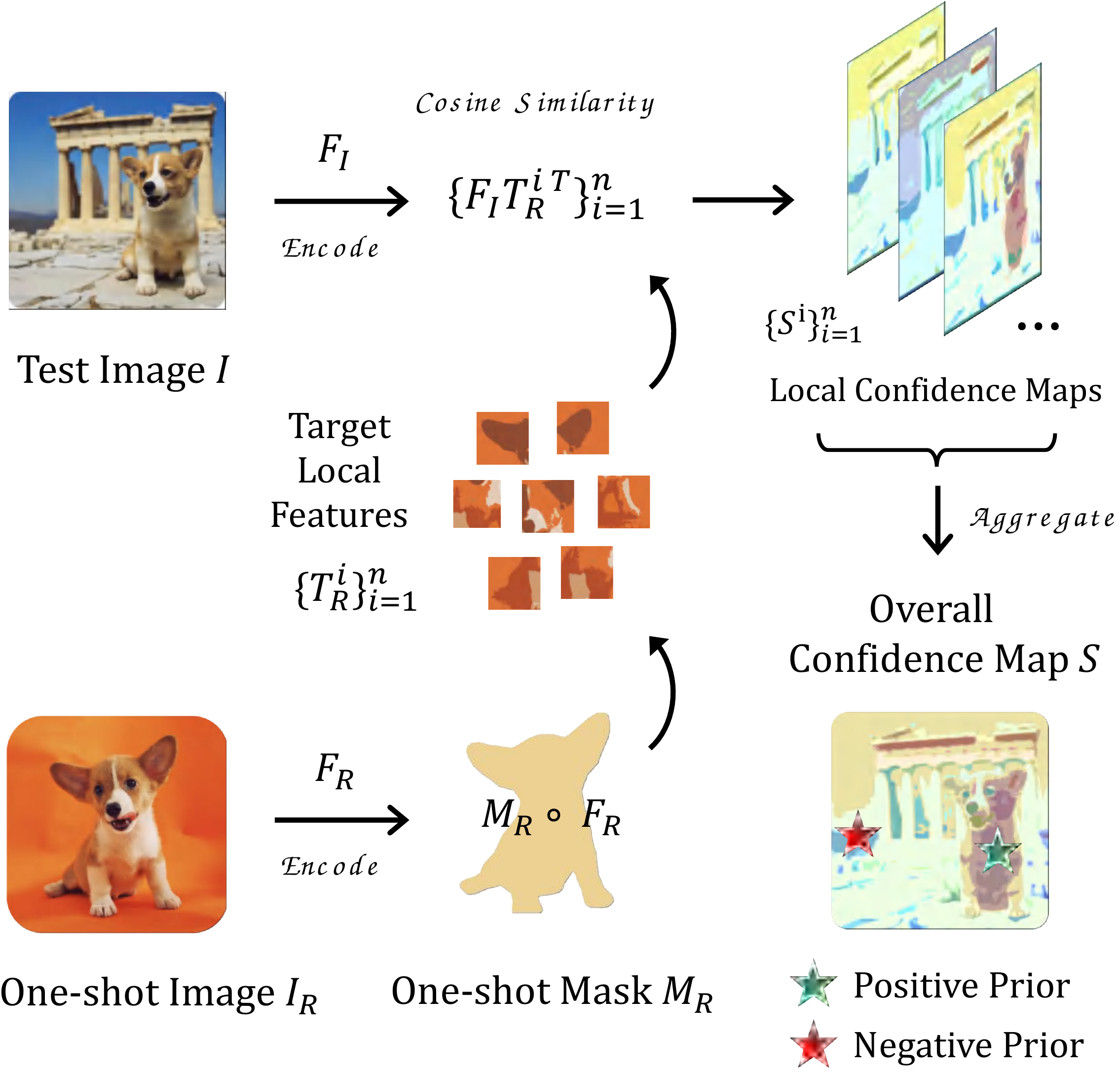}
\figcaption{\textbf{Positive-negative Location Prior.} We calculate a location confidence map for the target object in new test image by the appearance of all local parts. Then, we select the location prior as the point prompt for PerSAM.}
\label{fig4}
\end{minipage}
\hspace{0.2in}
\begin{minipage}[t]{0.47\textwidth}
\includegraphics[width=\textwidth]{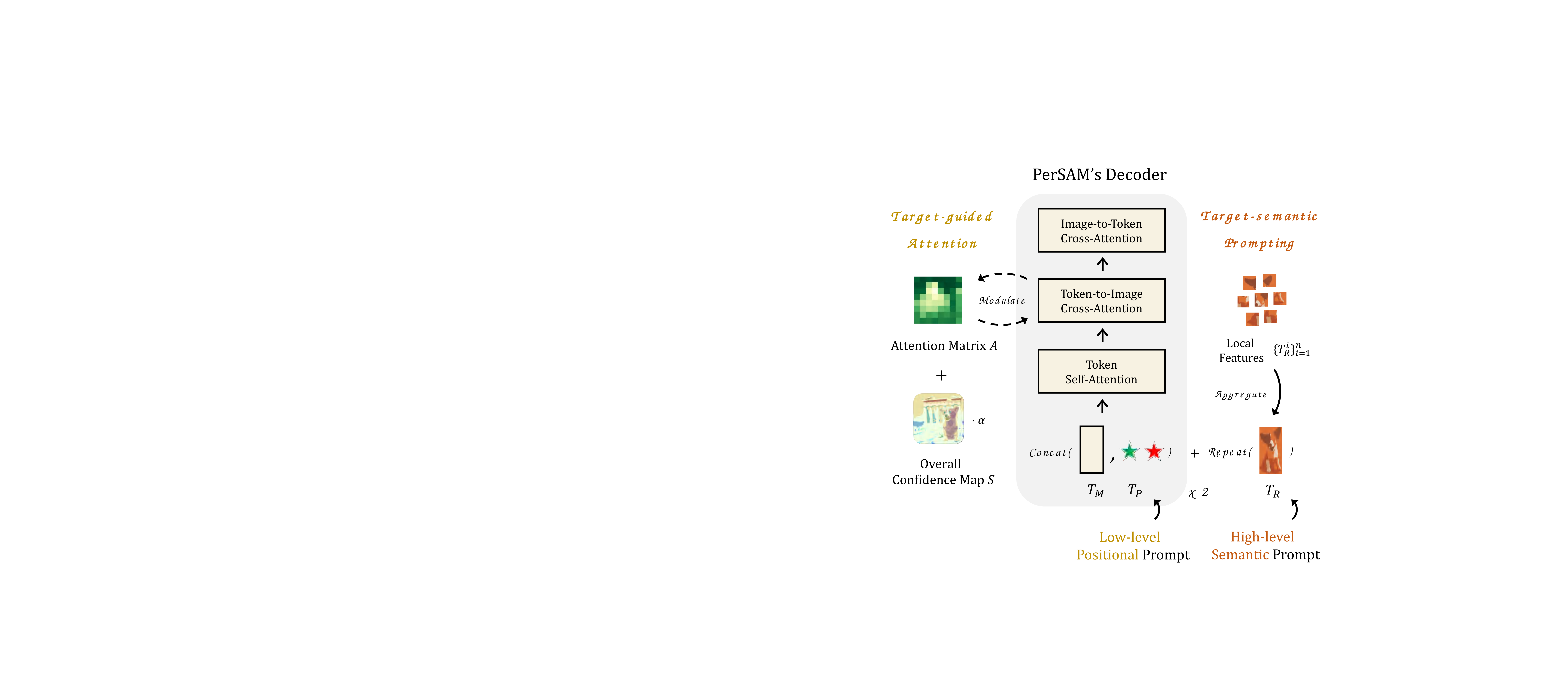}
\figcaption{\textbf{Target-guided Attention (Left) \& Target-semantic Prompting (Right).} To inject SAM with target semantics, we explicitly guide the cross-attention layers, and propose additional prompting with high-level cues.}
\label{fig5}
\end{minipage}
\end{figure*}

\subsection{Training-free PerSAM}
\label{s3.2}

\paragraph{Location Confidence Map.}
Conditioned on the user-provided image $I_R$ and mask $M_R$, PerSAM first obtains a confidence map that indicates the location of the target object in the new test image $I$.
As shown in Figure~\ref{fig4}, we apply an image encoder to extract the visual features of both $I_R$ and $I$. The encoder can be SAM's frozen backbone or other pre-trained vision models, for which we adopt SAM's image encoder $\operatorname{Enc}_I$ by default. We formulate the process as
\begin{align}
    F_I = \operatorname{Enc}_I(I),\ \ \ F_R = \operatorname{Enc}_I(I_R),
\end{align}
where $F_I, F_R \in \mathbb{R}^{h\times w\times c}$.
Then, we utilize the reference mask $M_R \in \mathbb{R}^{h\times w\times 1}$ to crop the features of foreground pixels within the visual concept from $F_R$, resulting in a set of $n$ local features as
\begin{align}
    \{T_R^i\}_{i=1}^n = M_R \circ F_R,
\end{align}
where $T_R^i \in \mathbb{R}^{1\times c}$ and $\circ$ denotes spatial-wise multiplication. After this, we calculate $n$ confidence maps for each foreground pixel $i$ by the cosine similarity between $T_R^i$ and test image feature $F_I$ as
\begin{align}
    \{S^i\}_{i=1}^n = \{F_I T_R^{i\ T}\}_{i=1}^n,\quad \text{where}\ S^i \in \mathbb{R}^{h\times w}.
\end{align}
Note that $F_I$ and $T_R^i$ have been pixel-wisely L2-normalized. Each $S^i$ represents the distribution probability for a different local part of object in the test image, such as the head, the body, or the paws of a dog. On top of this, we adopt an average pooling to aggregate all $n$ local maps to obtain the overall confidence map of the target object as 
\begin{align}
    S = \frac{1}{n}\sum_{i=1}^nS^i\ \in \mathbb{R}^{h\times w}.
\label{sim}
\end{align}
By incorporating the confidences of every foreground pixel, $S$ can take the visual appearance of different object parts into consideration, and acquire a relatively comprehensive location estimation.

\paragraph{Positive-negative Location Prior.}
To provide PerSAM with a location prior on the test image, we select two points with the highest and lowest confidence values in $S$, denoted as $P_h$ and $P_l$, respectively. The former represents the most likely center position of the target object, while the latter inversely indicates the background. Then, they are regarded as the positive and negative point prompts, and fed into the prompt encoder as
\begin{align}
    T_P = \operatorname{Enc}_P(P_h, P_l)\ \in \mathbb{R}^{2\times c},
\end{align}
which denote the prompt tokens for SAM's decoder.
In this way, SAM would tend to segment the contiguous region surrounding the positive point, while discarding the negative one's on the image.

\paragraph{Target-guided Attention.}
Although the positive-negative point prompt has been obtained, we further propose a more explicit semantic guidance to the cross-attention operation in SAM's decoder, which concentrates the feature aggregation within foreground target regions.
As shown in Figure~\ref{fig5}, the overall confidence map $S$ in Equation~\ref{sim} can clearly indicate the rough region of the target visual concept in the test image (hotter colors indicate higher scores). Based on such a property, we utilize $S$ to guide the attention map in every token-to-image cross-attention layer of the decoder. Specifically, we denote every attention map after the $\operatorname{softmax}$ function as $A\in \mathbb{R}^{h\times w}$, and then modulate its attention distribution by
\begin{align}
    A^g = \operatorname{softmax}\Big(A + \alpha\cdot\operatorname{softmax}(S)\Big),
\label{alpha}
\end{align}
where $\alpha$ denotes a balancing factor. With the attention bias, the mask and prompt tokens are compelled to capture more visual semantics associated with the target subject, other than the unimportant background area.
This contributes to more effective feature aggregation in attention mechanisms, and enhances the final segmentation accuracy of PerSAM in a training-free manner.

\paragraph{Target-semantic Prompting.}
The vanilla SAM only receives prompts with low-level positional information, such as the coordinate of a point or a box. To provide SAM's decoder with more high-level cues, we propose to utilize the visual feature of the target concept as an additional high-level semantic prompting. We first obtain the global embedding $T_R$ of the object in the reference image by both average pooling between different local features as
\begin{align}
    T_R = \frac{1}{n}\sum_{i=1}^nT_R^i\ \in \mathbb{R}^{1\times c}.
\end{align}
Then, we element-wisely add $T_R$ to all the input tokens of the test image in Equation~\ref{input_tokens}, before feeding them into the decoder block, which is shown in Figure~\ref{fig5} as
\begin{align}
    T^g = \operatorname{Repeat}(T_R) + \operatorname{Concat}(T_M, T_P),
\end{align}
where $T^g$ denotes the input token guided by target semantics for the decoder $\operatorname{Dec}_M$, and the $\operatorname{Repeat}$ operation duplicates the target visual embedding.
Aided by the simple token incorporation, PerSAM is not only prompted by low-level location points, but also high-level target visual cues.

\begin{figure*}[t!]
\vspace{-0.2cm}
\begin{minipage}[t]{0.45\textwidth}
\includegraphics[width=\textwidth]{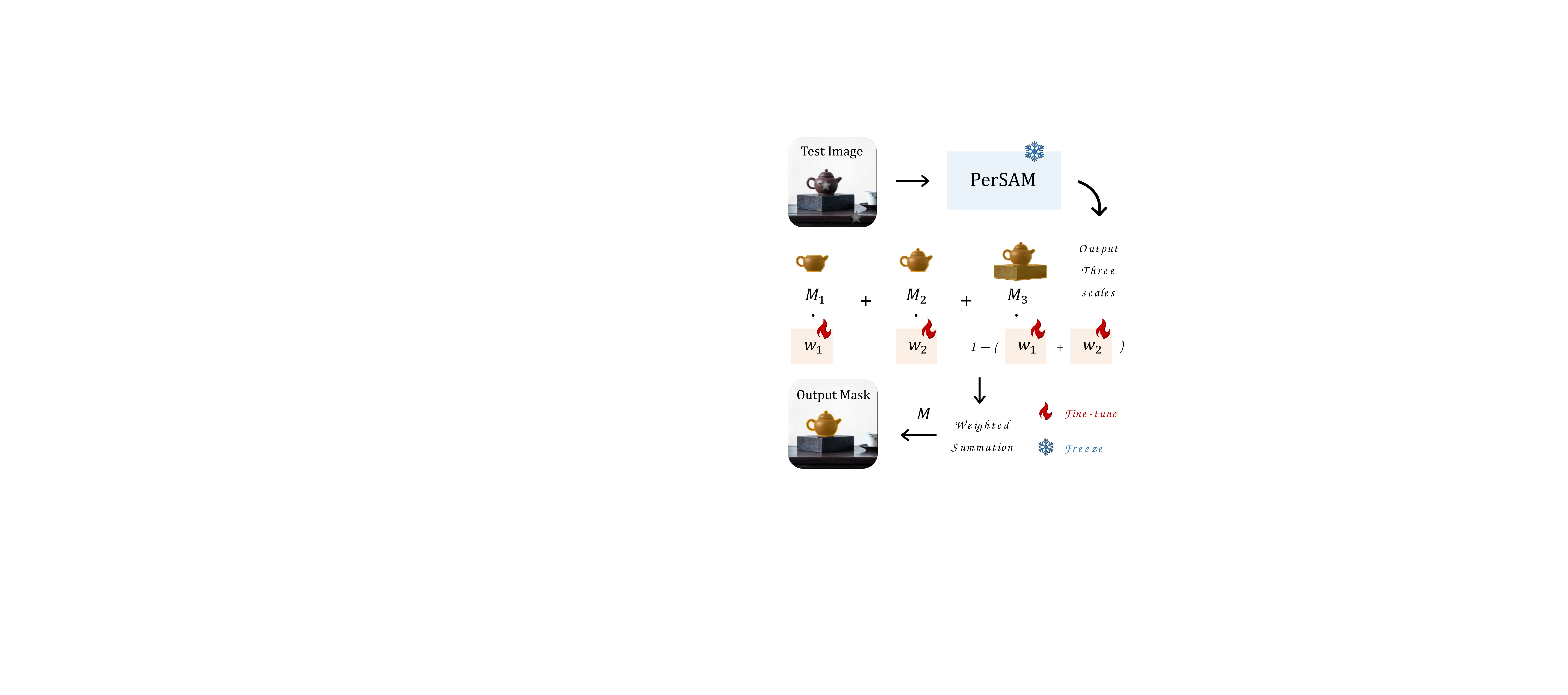}
\figcaption{\textbf{The Scale-aware Fine-tuning in PerSAM-F.} To alleviate the scale ambiguity, PerSAM-F adopts two learnable weights for adaptively aggregating three-scale masks.}
\label{fig6}
\end{minipage}
\hspace{0.27in}
\begin{minipage}[t]{0.45\textwidth}
\includegraphics[width=\textwidth]{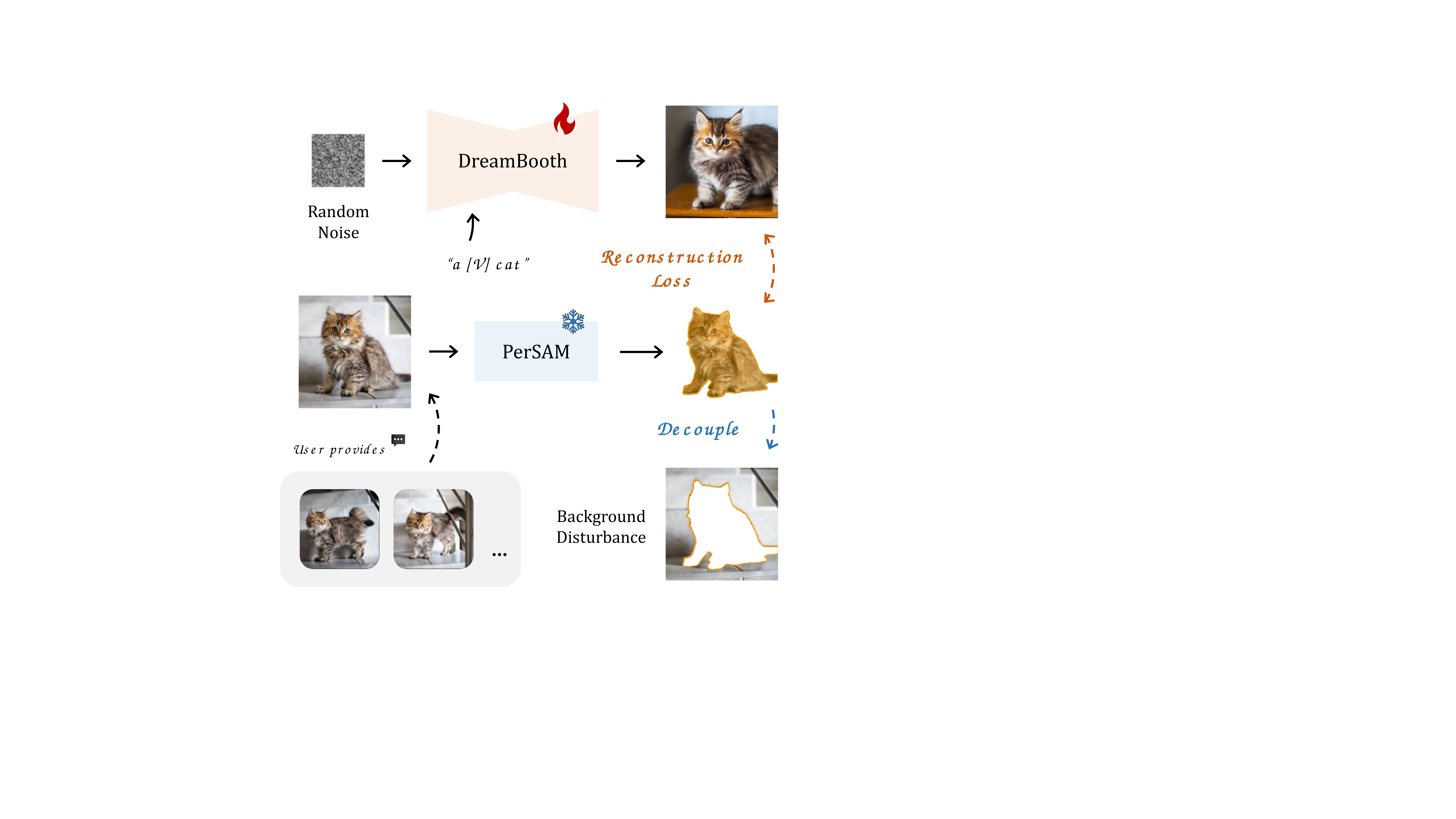}
\figcaption{\textbf{PerSAM-assisted DreamBooth.} We utilize PerSAM to decouple the target objects from the background for improving the generation of DreamBooth.}
\label{fig_db2}
\end{minipage}
\end{figure*}

\paragraph{Cascaded Post-refinement.}
Via the above techniques, we obtain an initial segmentation mask on the test image from SAM's decoder, which however, might include rough edges and isolated background noises. For further refinement, we iteratively feed the mask back into the decoder $\operatorname{Dec}_M$ for a two-step post-processing.
In the first step, we prompt the decoder by the currently predicted mask along with the previous positive-negative point prompt. For the second step, we acquire the bounding box enclosing the mask from the first step, and prompt the decoder additionally with this box for more accurate object localization.
As we only iterate the lightweight decoder without the large-scale image encoder, the post-processing is efficient and only costs an extra 2\% latency.

\subsection{Fine-tuning of PerSAM-F}
\label{s3.3}

\paragraph{Ambiguity of Segmentation Scales.}
The training-free PerSAM can tackle most cases with satisfactory segmentation accuracy. However, some target objects contain hierarchical structures, which leads to the ambiguity of mask scales. As shown in Figure~\ref{fig6}, the teapot on top of a platform is comprised of two parts: a lid and a body. If the positive point prompt (denoted by a green pentagram) is located at the body, while the negative prompt (denoted by a red pentagram) does not exclude the platform in a similar color, PerSAM would be misled for segmentation.
Such an issue is also discussed in SAM~\citep{kirillov2023segment}, where it proposes an alternative to simultaneously generate multiple masks of three scales, corresponding to the whole, part, and subpart of an object. Then, the user is required to manually select one mask out of three, which is effective but consumes extra manpower. In contrast, our personalized task aims to customize SAM for automatic object segmentation without the need for human prompting. This motivates us to further develop a scale-aware version of PerSAM by parameter-efficient fine-tuning.

\paragraph{Scale-aware Fine-tuning.}
For adaptive segmentation with the appropriate scale, we introduce a fine-tuning variant, PerSAM-F. Unlike the training-free model only producing one mask, PerSAM-F first follows PerSAM to obtain the location prior, and refers to SAM's original solution to output three-scale masks, denoted as $M_1$, $M_2$, and $M_3$, respectively. On top of this, we adopt two learnable mask weights, $w_1, w_2$, and calculate the final mask output by a weighted summation as
\begin{align}
    M = w_1\cdot M_1 + w_2\cdot M_2 + (1-w_1-w_2)\cdot M_3,
\end{align}
where $w_1, w_2$ are both initialized as $1/3$. To learn the optimal weights, we conduct one-shot fine-tuning on the reference image, and regard the given mask as the ground truth.
Note that, we freeze the entire SAM model to preserve its pre-trained knowledge, and only fine-tune the \textbf{\textit{2 parameters}} of $w_1, w_2$ within \textbf{\textit{10 seconds}} on a single A100 GPU.
In this way, our PerSAM-F efficiently learns the scale-aware semantics of objects, and adaptively outputs the best segmentation scale for different concepts, improving the generalization capacity of PerSAM.

\subsection{PerSAM-assisted DreamBooth}
\label{s3.4}

For personalized text-to-image synthesis, DreamBooth~\citep{ruiz2022dreambooth} fine-tunes a pre-trained diffusion model by the given 3$\sim$5 photos of a specific object, i.g., a pet cat. It learns to generate the cat referred to by a text prompt, ``a [V] cat'', and calculates the loss over the entire reconstructed images. This 
This would inject the redundant background information in the training images into the identifier [V].
Therefore, as shown in Figure~\ref{fig_db2}, we introduce our strategy to alleviate the disturbance of backgrounds in DreamBooth.
Given an object mask for any of the few-shot images, we leverage our PerSAM to segment all the foreground targets, and discard the gradient back-propagation for pixels belonging to the background area. Then, the Stable Diffusion is only fine-tuned to memorize the visual appearances of the target object. With no supervision imposed on the background, our PerSAM-assisted DreamBooth can not only synthesize the target object with better visual correspondence, but also increase the diversity of the new backgrounds guided by the input text prompt.

\begin{table*}[t!]
\small
\vspace{-0.2cm}
\centering
\caption{\textbf{Personalized Object Segmentation on the PerSeg Dataset}. We compare the overall mIoU, bIoU, and learnable parameters for different methods~\citep{bar2022visual,wang2022images,wang2023seggpt,zou2023segment}, along with the mIoU for 10 objects in PerSeg. `*' denotes works concurrent to ours.}
\begin{adjustbox}{width=\linewidth}
	\begin{tabular}{l|c c c| cc c c c c c c c c c}
	\toprule
  \makecell*[c]{Method} &\makecell*[c]{mIoU} &\makecell*[c]{bIoU} &\makecell*[c]{Param.}
        &\makecell*[c]{Can}  &\makecell*[c]{Barn}  &\makecell*[c]{Clock} &\makecell*[c]{Cat}
        &\makecell*[c]{Back-\\pack}
        &\makecell*[c]{Teddy\\Bear}&\makecell*[c]{Duck\\Toy}&\makecell*[c]{Thin\\Bird}&\makecell*[c]{Red\\Cartoon} &\makecell*[c]{Robot\\Toy}\\
		\cmidrule(lr){1-1} \cmidrule(lr){2-3} \cmidrule(lr){4-4} 
  \cmidrule(lr){5-14}
        Painter &56.4& 42.0 &354M &19.1 & 3.2 & 42.9 & 94.1 & 88.1 & 93.0 & 33.3 & 20.9 &98.2 & 65.0 \\
        VP &65.9 & 25.5 &383M & 61.2 & 58.6 & 59.2 & 76.6 & 66.7 & 79.8 & 89.9 & 67.4 & 81.0 & 72.4 \\
        SEEM* &87.1& 55.7 &341M & 65.4 & 82.5 & 72.4 & 91.1 & 94.1 & 95.2 & 98.0 & 71.3 & 97.0 & 95.8\\
        SegGPT* &94.3& 76.5 &354M &96.6 &63.8 &92.6 &94.1 &94.4 &93.7 &97.2 &92.6 &97.3 &96.2 \\
        \cmidrule(lr){1-14}
        \rowcolor{pink!12}  \bf PerSAM &89.3 &71.7 &0 &96.2 &38.9 &96.2 &90.70 &95.39 &94.6 &97.3 &93.7 &97.0 &60.6 \\
        \rowcolor{pink!12} \bf PerSAM-F &95.3 &77.9 &2 &96.7 &97.5 &96.1 &92.3 &95.5 &95.2 &97.3 &94.0 &97.1 &96.7\\
	\bottomrule
	\end{tabular}
\end{adjustbox}
 \label{t1}
\end{table*}

\begin{figure*}
\begin{minipage}[t!]{0.32\linewidth}
\centering
\tabcaption{\textbf{Video Object Segmentation} on DAVIS 2017 val~\citep{pont20172017}. We utilize gray color to denote the methods involving in-domain training.}
\label{t2}
\begin{adjustbox}{width=\linewidth}
	\begin{tabular}{lccc}
	\toprule
		\makecell*[l]{Method} &$\mathcal{J}\&\mathcal{F}$ & $\mathcal{J}$ &$\mathcal{F}$\\
		\cmidrule(lr){1-1} \cmidrule(lr){2-4}
        \color{gray}{AGSS}&\color{gray}{67.4} &\color{gray}{64.9} &\color{gray}{69.9}\\
        \color{gray}{AFB-URR}&\color{gray}{74.6} &\color{gray}{73.0} &\color{gray}{76.1}\\
      \cmidrule(lr){1-4}
        Painter&34.6 &28.5 &40.8\\
        SEEM&58.9 &55.0 &62.8\\
        SegGPT&75.6 &72.5 &78.6\vspace{0.05cm}\\
        \rowcolor{pink!12}\bf PerSAM &66.9 &71.3 &75.1\\
        \rowcolor{pink!12}\bf PerSAM-F&76.1 &74.9 &79.7\\
	  \bottomrule
	\end{tabular}
\end{adjustbox}
\end{minipage}\qquad
\begin{minipage}[t!]{0.63\linewidth}
\centering
\tabcaption{\textbf{One-shot Semantic and Part Segmentation} on FSS-1000~\citep{li2020fss}, LVIS-92$^{i}$~\citep{gupta2019lvis}, PASCAL-Part~\citep{morabia2020attention}, and PACO-Part~\citep{ramanathan2023paco}. We report the mIoU scores and utilize gray color to denote the methods involving in-domain training.}
\label{t3}
\begin{adjustbox}{width=\linewidth}
\centering
	\begin{tabular}{lcccc}
	\toprule
 \multirow{2}*{Method} &\multicolumn{2}{c}{\ \ One-shot Semantic Seg.\ \ } &\multicolumn{2}{c}{One-shot Part Seg.}\\
		&FSS-1000 &LVIS-92$^{i}$&PASCAL-Part&PACO-Part\\
		\cmidrule(lr){1-1} \cmidrule(lr){2-3} \cmidrule(lr){4-5}
      \color{gray}{HSNet} & \color{gray}{86.5} & \color{gray}{17.4} & \color{gray}{32.4} & \color{gray}{22.6} \\
      \color{gray}{VAT} & \color{gray}{90.3} & \color{gray}{18.5} & \color{gray}{33.6} & \color{gray}{23.5} \\
      \cmidrule(lr){1-5}
      Painter & 61.7 & 10.5 & 30.4 & 14.1 \\
      SegGPT & 85.6 & 18.6 & - & - \vspace{0.05cm}\\
     \rowcolor{pink!12} \bf PerSAM & 81.6 & 15.6 & 32.5 & 22.5\\
      \rowcolor{pink!12}\bf PerSAM-F & 86.3 & 18.4 & 32.9 & 22.7 \\
	  \bottomrule
	\end{tabular}
\end{adjustbox}
\end{minipage}
\end{figure*}

\section{Experiment}
\vspace{-0.1cm}

We first evaluate our approach for personalized segmentation on PerSeg in Section~\ref{s4.1}, along with various existing one-shot segmentation benchmarks in Section~\ref{s4.2}. Then, we illustrate the effectiveness of our PerSAM-assisted DreamBooth in Section~\ref{s4.3}.
Finally, we conduct several ablation studies to investigate our designs on PerSeg in Section~\ref{s4.4}.

\subsection{Personalized Evaluation}
\label{s4.1}

\paragraph{PerSeg Dataset.}
To test the personalization capacity, we construct a new segmentation dataset, termed PerSeg. The raw images are collected from the training data of subject-driven diffusion works~\citep{ruiz2022dreambooth,gal2022image,kumari2022multi}. PerSeg contains 40 objects of various categories in total, including daily necessities, animals, and buildings. In different poses or scenes, each object is associated with 5$\sim$7 images and masks, where we fix one image-mask pair as the user-provided one-shot data. The mIoU and bIoU~\citep{cheng2021boundary} are adopted for evaluation. Please refer to the Appendix for implementation details and an enlarged data scale of PerSeg.
\vspace{-0.1cm}

\paragraph{Performance.}
In Table~\ref{t1}, we observe the fine-tuned PerSAM-F achieves the best results, which effectively enhances PerSAM by +2.7\% and +5.9\% overall mIoU and bIoU. We show more visualization of PerSAM-F's improvement in Figure~\ref{fig_vis}.
Visual Prompting (VP)~\citep{bar2022visual}, Painter~\citep{wang2022images}, SEEM~\citep{zou2023segment}, and SegGPT~\citep{wang2023seggpt} are in-context learners that can also segment objects according to the given one-shot prompt data.
As shown, the training-free PerSAM can already achieve better performance than Painter, VP, and SEEM with different margins. By the efficient 2-parameter fine-tuning, our PerSAM-F further surpasses the powerful SegGPT by +2.4\% and +4.1\% overall mIoU and bIoU. Different from their motivations to develop segmentation generalists, our method is specially designed for personalized object segmentation, and exhibits much more efficiency in both time and computational resources.

\subsection{Existing Segmentation Benchmarks}
\label{s4.2}

\begin{figure*}[t!]
\vspace{-0.2cm}
\begin{minipage}[t]{0.47\textwidth}
\includegraphics[width=\textwidth]{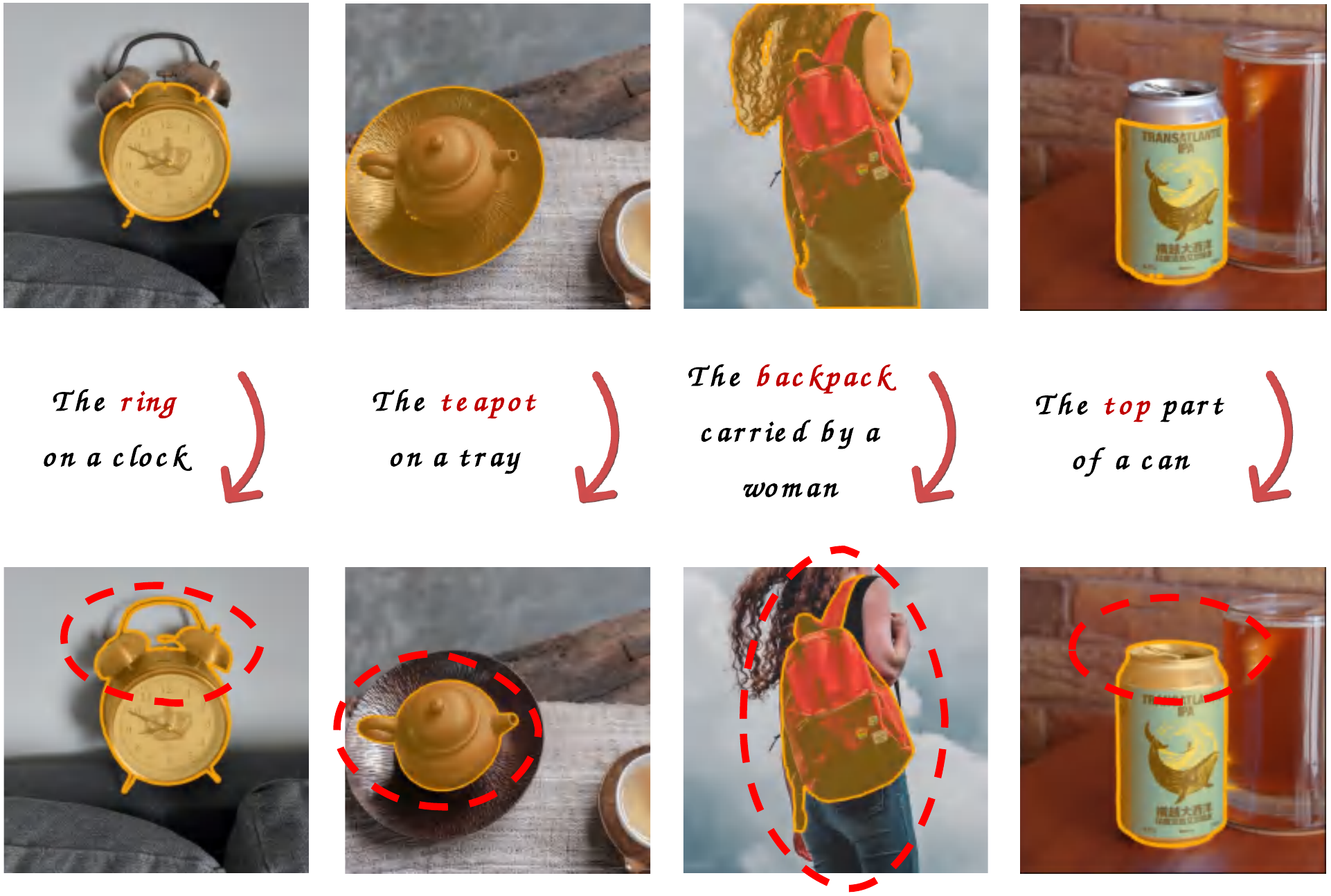}
\figcaption{\textbf{Visualization of PerSAM-F's Improvement.} Our scale-aware fine-tuning can well alleviate the scale ambiguity of PerSAM.}
\label{fig_vis}
\end{minipage}
\hspace{0.24in}
\vspace{-0.2cm}
\begin{minipage}[t]{0.47\textwidth}
\includegraphics[width=\textwidth]{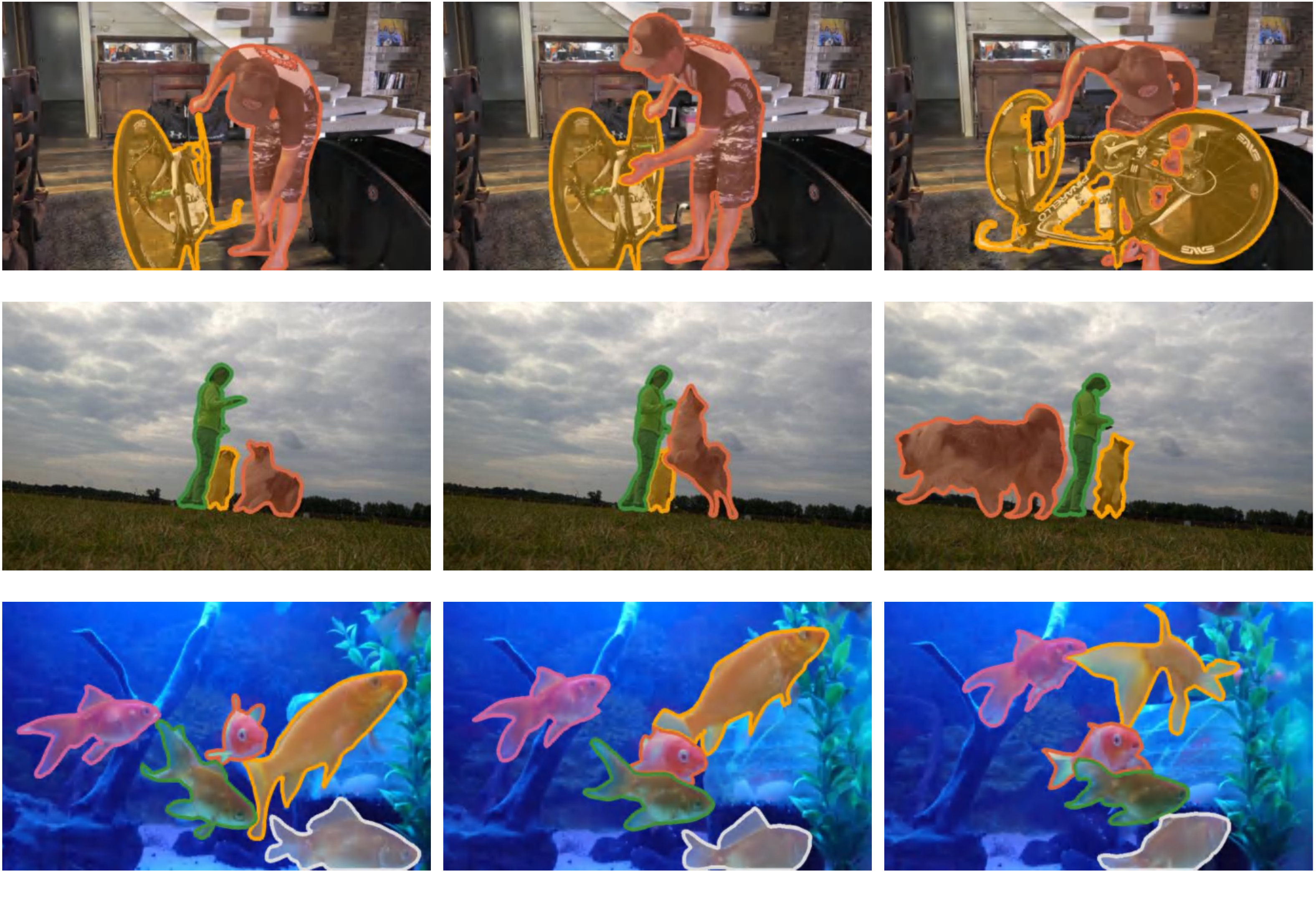}
\figcaption{\textbf{Visualization of Video Object Segmentation.} Our approach performs well for segmenting multiple objects in a video.}
\label{fig_video}
\end{minipage}
\end{figure*}

\paragraph{Video Object Segmentation.}
Given the first-frame image and object masks, our PerSAM and PerSAM-F achieve competitive object segmentation and tracking performance on the validation set of DAVIS 2017~\citep{pont20172017} 
As shown in Table~\ref{t2}, compared to methods without video training, the training-free PerSAM largely surpasses Painter by +32.3\% $\mathcal{J}\&\mathcal{F}$ score, and our PerSAM-F can achieve +0.5\% better performance than SegGPT. Notably, our one-shot fine-tuning approach can outperform methods~\citep{lin2019agss,liang2020video} fully trained by extensive video data. The results fully illustrate our strong generalization ability for temporal video data and complex scenarios, which contain multiple similar or occluded objects, as visualized in Figure~\ref{fig_video}.
\vspace{-0.1cm}

\paragraph{One-shot Semantic and Part Segmentation.}
In Table~\ref{t3}, we evaluate our approach for one-shot image segmentation respectively on four datasets, FSS-1000~\citep{li2020fss}, LVIS-92$^{i}$~\citep{gupta2019lvis}, PASCAL-Part~\citep{morabia2020attention}, and PACO-Part~\citep{ramanathan2023paco}, where we follow Matcher~\citep{liu2023matcher} for data pre-processing and evaluation. As shown, our PerSAM-F attains consistently better results than Painter, and performs comparably to SegGPT. For models~\citep{min2021hypercorrelation,hong2022cost} with in-domain training, our approach can achieve higher scores than HSNet. The experiments well demonstrate that, our proposed approach is not limited to object-level segmentation, but also works for category-wise and part-wise personalization of SAM.

\subsection{PerSAM-assisted DreamBooth}
\label{s4.3}

\begin{figure*}[t]
  \centering
  \vspace{-0.2cm}
\includegraphics[width=\textwidth]{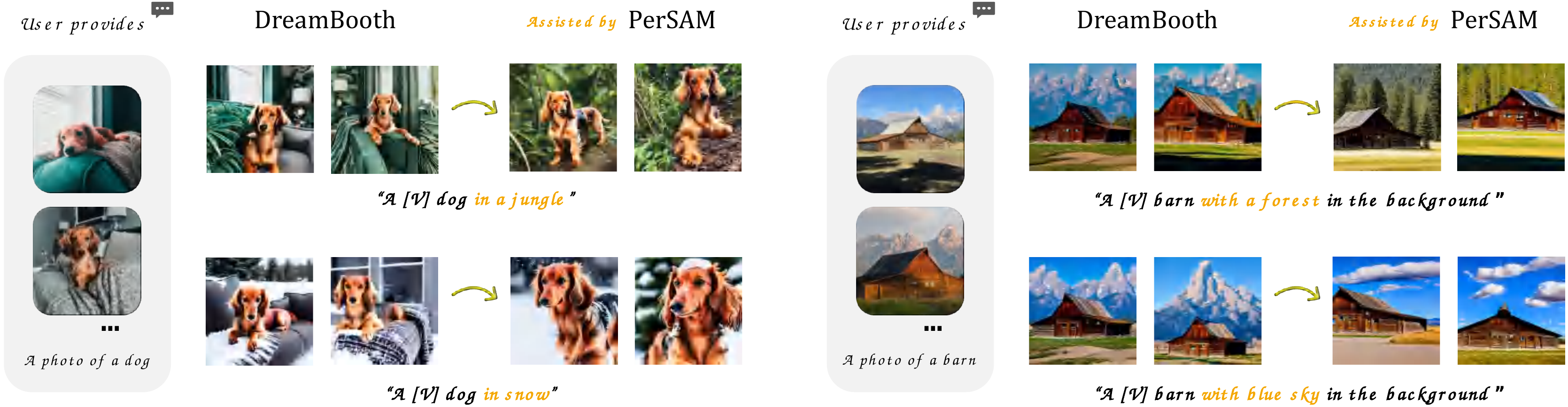}
   \caption{\textbf{Visualization of PerSAM-guided DreamBooth.} The improved DreamBooth~\citep{ruiz2022dreambooth} can better preserve the diversity for synthesizing various contexts in new images.}
    \label{fig_db3}
\end{figure*}

We follow all the hyperparameters in DreamBooth~\citep{ruiz2022dreambooth} to fine-tune a pre-trained Stable Diffusion~\citep{rombach2022high} for personalized image synthesis.
In addition to Figure~\ref{fig_db1}, we visualize more examples of PerSAM-assisted DreamBooth in Figure~\ref{fig_db3}. For the dog lying on a grey sofa, the ``jungle'' and ``snow'' by DreamBooth are still the sofa with green and white decorations. Assisted by PerSAM-F, the newly-generated background is totally decoupled with the sofa and well corresponds to the textual prompt. For the barn in front of the mountains, our approach also alleviates the background disturbance to correctly generate the ``forest'' and ``blue sky''.

\subsection{Ablation Study}
\label{s4.4}

\textbf{Main Components.} In Table~\ref{t4}, we investigate our different components by starting from a baseline that only adopts the positive location prior. Then, we add the negative point prompt and cascaded post-refinement, enhancing +3.6\% and +11.4\% mIoU, respectively.
On top of that, we introduce the high-level target semantics into SAM's decoder for attention guidance and semantic prompting. The resulting +1.9\% and +3.5\% improvements fully indicate their significance. Finally, via the efficient scale-aware fine-tuning, PerSAM-F boosts the score by +6.0\%, demonstrating superior accuracy.

\textbf{Different Fine-tuning Methods.}
In Table~\ref{t5}, we experiment with other parameter-efficient fine-tuning (PEFT) methods for PerSAM-F, i.e., prompt tuning~\citep{liu2021p-tuning}, Adapter~\citep{houlsby2019parameter}, and LoRA~\citep{hu2021lora}. We freeze the entire SAM, and only tune the PEFT modules injected into every transformer block in PerSAM's decoder.
As shown, the prompt tuning and Adapter would over-fit the one-shot data and severely degrade the accuracy. Instead, our scale-aware fine-tuning can best improve the performance of PerSAM, while tuning the least learnable parameters.

\textbf{Using Box-image as Reference.}
Requiring an accurate mask as one-shot data might be too strict for some users.
In Table~\ref{t6}, we relax the input restrictions to a bounding box designating the expected object.
For our method, we can regard the box as a prompt and utilize off-the-shelf SAM to generate the one-shot mask. Therefore, the box reference only leads to a marginal performance drop in PerSAM and PerSAM-F, but severely influences other methods.
\vspace{-0.1cm}

\section{Conclusion}

In this paper, we propose to personalize Segment Anything Model (SAM) for specific visual concepts with only one-shot data. Firstly, we introduce PerSAM, which injects high-level target semantics into SAM with training-free techniques. On top of this, we present a scale-aware fine-tuning variant, PerSAM-F. With only 2 learnable parameters, PerSAM-F effectively alleviates the ambiguity of mask scales and achieves leading performance on various benchmarks. Besides, we also verify the efficacy of our approach to assist DreamBooth in fine-tuning better text-to-image diffusion models. 
We hope our work may expand the applicability of SAM to a wider range of scenarios.

\begin{figure*}
\begin{minipage}[t!]{0.34\linewidth}
\centering
 \small
\tabcaption{\textbf{Ablation of Main Components} in our proposed method.}
\label{t4}
\begin{adjustbox}{width=\linewidth}
\centering
	\begin{tabular}{lcl}
	\toprule
		Variant &mIoU &Gain\\
		\cmidrule(lr){1-1} \cmidrule(lr){2-2} \cmidrule(lr){3-3}
      Positive Prior &69.1 &- \\
      \cmidrule(lr){1-3}
      + Negative Prior &72.5 &\color{blue}{+3.4}  \\
      + Post-refinement &83.9 &\color{blue}{+11.4} \\
      + Guided Attention &85.8 &\color{blue}{+1.9} \\
      + Semantic Prompt &89.3 &\color{blue}{+3.5} \\
      \cmidrule(lr){1-3}
      + Scale Tuning &95.3 &\color{blue}{+6.0} \\
	  \bottomrule
	\end{tabular}
\end{adjustbox}
\end{minipage}
\hspace{0.2cm}
\begin{minipage}[t!]{0.325\linewidth}
\centering
 \small
\tabcaption{\textbf{Ablation of Different Fine-tuning Methods}.}
\label{t5}
\begin{adjustbox}{width=\linewidth}
\centering
	\begin{tabular}{llc}
	\toprule
		Method &Param. &mIoU \\
		\cmidrule(lr){1-1} \cmidrule(lr){2-2} \cmidrule(lr){3-3} 
      PerSAM &0 &89.32  \\
      \cmidrule(lr){1-3}
      Prompt Tuning  &12K &76.5\\
      Adapter  &196K &78.3 \\
      LoRA & 293K & 90.0 \\
      3 Mask Weights &3 &92.9\\
      \cmidrule(lr){1-3}
      PerSAM-F &2 &95.3 \\
	  \bottomrule
	\end{tabular}
\end{adjustbox}
\end{minipage}
\hspace{0.2cm}
\begin{minipage}[t!]{0.275\linewidth}
\centering
 \small
\tabcaption{\textbf{Ablation of using Box-image as Reference}.}
\label{t6}
\begin{adjustbox}{width=\linewidth}
\centering
	\begin{tabular}{llc}
	\toprule
      Method &Mask &Box \\
		\cmidrule(lr){1-1} \cmidrule(lr){2-2} \cmidrule(lr){3-3} 
      Painter &56.4	&42.0  \\
      VP &65.9	&38.1  \\
      SEEM &87.1	&64.9  \\
      SegGPT  &94.3	&36.0\\
      \cmidrule(lr){1-3}
      PerSAM &89.3	&88.1 \\
      PerSAM-F &95.3	&94.9\\
	  \bottomrule
	\end{tabular}
\end{adjustbox}
\end{minipage}
\end{figure*}


\clearpage
\bibliography{iclr2024_conference}
\bibliographystyle{iclr2024_conference}

\end{document}